%% file: icml2024.tex
\documentclass{article}

\usepackage{microtype}
\usepackage{graphicx}
\usepackage{subfigure}
\usepackage{booktabs} % for professional tables
\usepackage{lscape}
\usepackage{listings}
\usepackage{multicol}

\usepackage{hyperref}

\newcommand{\mat}[1]{\mathbf{#1}}
\usepackage[accepted]{icml2024preprint} % for 

\usepackage{amsmath}
\usepackage{amssymb}
\usepackage{mathtools}
\usepackage{amsthm}
\usepackage{comment}

\usepackage[capitalize,noabbrev]{cleveref}

\theoremstyle{plain}

\theoremstyle{definition}

\theoremstyle{remark}

\def\ie{{\frenchspacing\it i.e.}}
\def\eg{{\frenchspacing\it e.g.}}
\def\etc{{\frenchspacing\it etc.}}

\def\beq#1{\begin{equation}\label{#1}}
\def\eeq{\end{equation}}
\def\beqa#1{\begin{eqnarray}\label{#1}}
\def\eeqa{\end{eqnarray}}

\def\fig#1{Fig.~\ref{#1}}
\def\Fig#1{Figure~\ref{#1}}

\def\nhidden{n}
\def\fwidth{w_f}
\def\fdepth{d_f}
\def\gwidth{w_g}
\def\gdepth{d_g}
\def\hyperpars{{\bf p}}

\def\k{{\bf k}}
\def\x{{\bf x}}
\def\y{{\bf y}}
\def\h{{\bf h}}
\def\A{{\bf A}}

\usepackage[textsize=tiny]{todonotes}

\icmltitlerunning{
Program synthesis via mechanistic interpretability}

\usepackage{tikz}
\usetikzlibrary{shapes.geometric, arrows.meta, positioning}

\begin{document}

\twocolumn[
\icmltitle{Opening the AI black box:\\
program synthesis via mechanistic interpretability}



\icmlsetsymbol{equal}{*}

\begin{icmlauthorlist}
% Some permutation of Ziming Liu, Eric Michaud, Isaac Liao, Vedang Lad, Carl Guo, Tara Rezaei Kheirkhah, Anish Mudide, Chloe Loughridge, Mateja Vukeli\'c, and Max Tegmark
% Put Armaun & James here or in acknowledgements
\icmlauthor{Eric J. Michaud}{equal,mit,iaifi}
\icmlauthor{Isaac Liao}{equal,mit}
\icmlauthor{Vedang Lad}{equal,mit}
\icmlauthor{Ziming Liu}{equal,mit,iaifi}
\icmlauthor{Anish Mudide}{mit}
\icmlauthor{Chloe Loughridge}{harvard}
\icmlauthor{Zifan Carl Guo}{mit}
% \icmlauthor{}{sch}
\icmlauthor{Tara Rezaei Kheirkhah}{mit}
\icmlauthor{Mateja Vukeli\'c}{mit}
\icmlauthor{Max Tegmark}{mit,iaifi}
\end{icmlauthorlist}

\icmlaffiliation{mit}{Massachusetts Institute of Technology, Cambridge, MA, USA}
\icmlaffiliation{iaifi}{Institute for Artificial Intelligence and Fundamental Interactions}
\icmlaffiliation{harvard}{Harvard University, Cambridge, MA, USA}


\icmlcorrespondingauthor{Max Tegmark}{tegmark@mit.edu}

\icmlkeywords{Mechanistic Interpretability, Formal Verification}\vskip 0.3in]



\printAffiliationsAndNotice{\icmlEqualContribution} % otherwise use the standard text.

\begin{abstract}
We present MIPS, a novel method for program synthesis based on automated mechanistic interpretability of neural networks trained to perform the desired task, auto-distilling the learned algorithm into Python code.
We test MIPS on a benchmark of 62 algorithmic tasks that can be learned by an RNN and find it highly complementary to GPT-4: MIPS solves 32 of them, including 13 that are not solved by GPT-4 (which also solves 30). 
MIPS uses an integer autoencoder to convert the RNN into a finite state machine, then applies Boolean or integer symbolic regression to capture the learned algorithm.
As opposed to large language models, this program synthesis technique makes no use of (and is therefore not limited by) human training data such as algorithms and code from GitHub.
We discuss opportunities and challenges for scaling up this approach to make machine-learned models more interpretable and trustworthy.
\end{abstract}

\section{Introduction}

Machine-learned algorithms now outperform traditional human-discovered algorithms on many tasks, from translation to general-purpose reasoning. These learned algorithms tend to be black-box neural networks, and we typically lack a full understanding of how they work.
This has motivated the growing field of {\it mechanistic interpretability}, aiming to assess and improve their trustworthiness.
Major progress has been made in interpreting and understanding smaller models, but this work has involved human effort, which raises questions about whether it can scale to larger models. This makes it timely to investigate whether mechanistic interpretability can be fully automated \cite{tegmark2023provably}. 

The goal of the present paper is to take a modest first step in this direction by presenting and testing MIPS ({\bf M}echanistic-{\bf I}nterpretability-based {\bf P}rogram {\bf S}ynthesis), a fully automated method that can distill simple learned algorithms from neural networks into Python code. The rest of this paper is organized as follows. After reviewing prior work in Section~II, we present our method in Section~III, test it on a benchmark in Section~IV and summarize our conclusions in Section~V.

\section{Related Work}

{\it Program synthesis} is a venerable field dating back to Alonzo Church in 1957; \citet{zhou2023survey} and \citet{odena2020bustle} provide recent reviews of the field. 
Large language Models (LLMs) have become increasingly good at writing code based on verbal problem descriptions or auto-complete.
We instead study the common alternative problem setting known as
``programming by example" (PBE), where the desired program is specified by giving examples of input-output pairs \cite{wu2023programming}. 
The aforementioned papers review a wide variety of program synthesis methods, many of which involve some form of search over a space of possible programs. LLMs that synthesize code directly have recently become quite competitive with such search-based approaches \cite{sobania2022choose}.
Our work provides an alternative search-free approach where the program learning happens during neural network training rather than execution.

Our work builds on the recent progress in {\it mechanistic interpretability} (MI) of neural networks~\citep{olah2020zoom, cammarata2020curve, wang2022interpretability, olsson2022context}.
Much MI work has tried to understand how neural networks represent various types of information, \eg, 
geographic information~\cite{goh2021multimodal, gurnee2023language},
truth~\cite{burns2022discovering, marks2023geometry} and the state of board games~\cite{mcgrath2022acquisition, toshniwal2022chess, li2022emergent}.
Another major MI thrust has been to understand how 
neural networks perform algorithmic tasks, \eg, modular arithmetic~\citep{nanda2023progress,liu2022towards,zhong2023clock,quirke2023understanding}, greater-than~\citep{hanna2023does}, and greatest-common-divisor~\cite{charton2023can}.

Whereas \citet{lindner2023tracr} automatically convert traditional code into a neural network, we aim to do the opposite. Other recent efforts to automate MI include 
identifying a sparse subgraph of the network whose units are causally relevant to a behavior of interest~\cite{conmy2023towards, syed2023attribution} 
and using LLMs to label internal components of neural networks, for instance neurons~\cite{bills2023language} and features discovered by sparse autoencoders~\cite{cunningham2023sparse, bricken2023monosemanticity}.

\section{MIPS, our program synthesis algorithm}

\begin{figure}[htbp]
    \centering
    \includegraphics[width=1\linewidth]{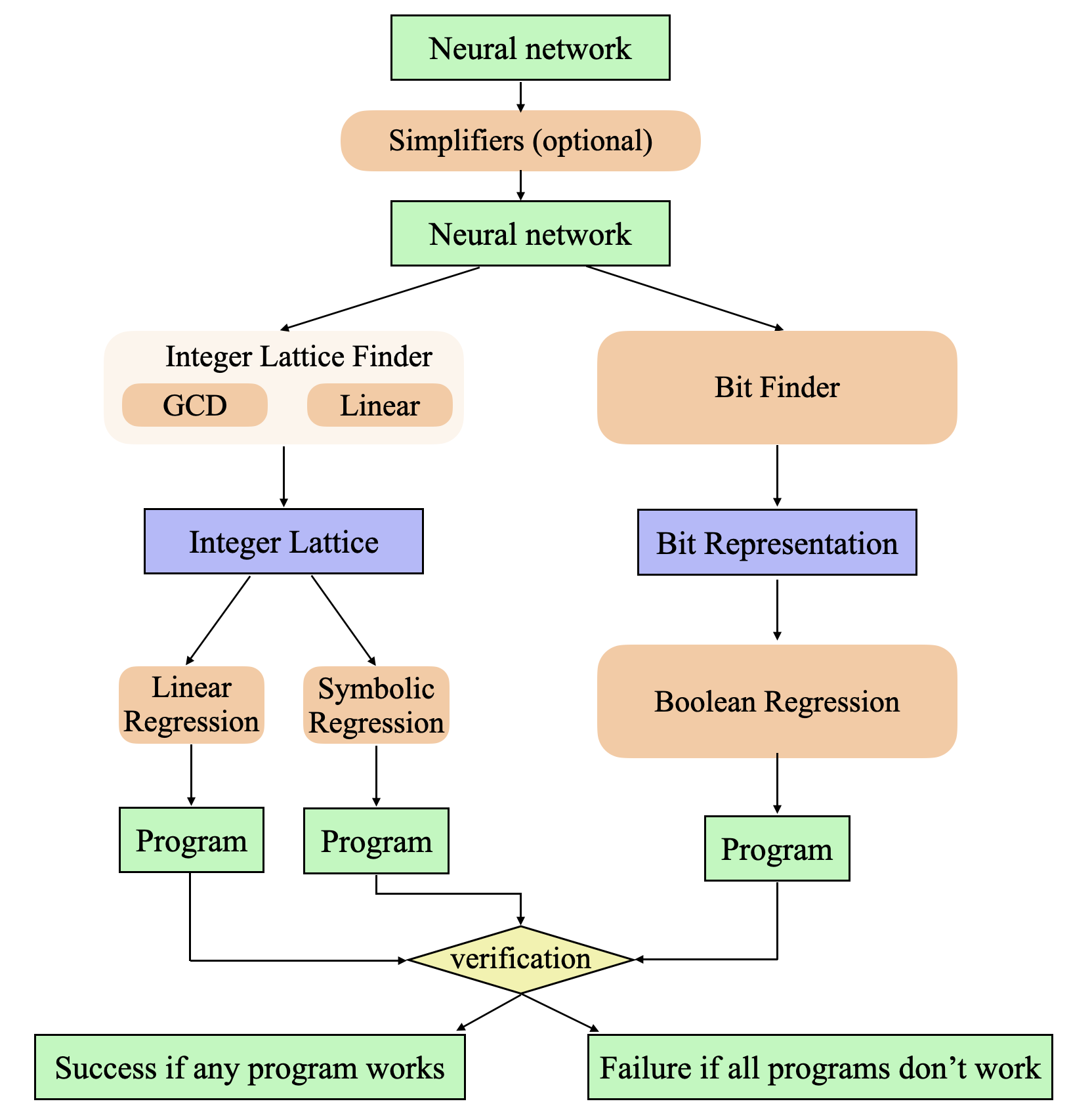}
    \caption{The pipeline of our program synthesis method. MIPS relies on discovering integer representations and bit representations of hidden states, which enable regression methods to figure out the exact symbolic relations between input, hidden, and output states.}
    \label{fig:program-synthesis-pipeline}
\end{figure}

As summarized in Figure~\ref{fig:program-synthesis-pipeline}, MIPS involves the following key steps.
\begin{enumerate}
\item Neural network training
\item Neural network simplification
\item Finite state machine extraction
\item Symbolic regression
\end{enumerate}
Step 1 is to train a black-box neural network to learn an algorithm that performs the desired task. 
In this paper, we use a recurrent neural network (RNN) of the general form
\begin{eqnarray}
\label{RNNeq}
\h_i&=&f(\h_{i-1},\x_i),\\
\y_i&=&g(\h_i),
\end{eqnarray}
that maps input vectors $\x_i$ into output vectors $\y_i$ via hidden states $\h_i$. The RNN is defined by the two functions $f$ and $g$, which are implemented as feed-forward neural networks (MLPs) to allow more model expressivity than for a vanilla RNN.
The techniques described below can also be applied to more general neural network architectures.

Step 2 attempts to automatically simplify the learned neural network without reducing its accuracy.
Steps 3 and 4 automatically distill this simplified learned algorithm into Python code.
When the training data is discrete (consisting of say text tokens, integers, or pixel colors), the neural network will be a {\it finite state machine}: the activation vectors for each of its neuron layers define finite sets and the entire working of the network can be defined by look-up tables specifying the update rules for each layer. For our RNN, this means that the space of hidden states $\h$ is discrete, so that the functions $f$ and $g$ can be defined by lookup tables.
As we will see below, the number of hidden states that MIPS needs to keep track of can often be greatly reduced by clustering them, corresponding to learned representations. After this, the geometry of the cluster centers in the hidden space often reveals 
that they form either an incomplete multidimensional lattice whose points represent integer tuples, or a set whose cardinality is a power of two, whose points represent Boolean tuples. 
In both of these cases, the aforementioned lookup tables simply specify integer or Boolean functions, which MIPS attempts to discover via symbolic regression. Below we present an {\it integer autoencoder} and a {\it Boolean autoencoder} to discover such integer/Boolean representations from arbitrary point sets.

We will now describe each of the three steps of MIPS in greater detail.

%\subsection{Neural architecture search and training}
\subsection{AutoML optimizing for simplicity}\label{sec:automl}

We wish to find the {\it simplest} RNN that can learn our task, to facilitate subsequent discovery of the algorithm that it has learned. We therefore implement an AutoML-style neural architecture search that tries to minimize network size
while achieving perfect test accuracy. This search space is defined by a vector $\hyperpars$ of five main architecture hyperparameters: the five integers
$\hyperpars = (\nhidden,\fwidth,\fdepth,\gwidth,\gdepth)$ corresponding to the 
dimensionality of hidden state $\h$, the width and 
depth of the $f$-network, and the width and 
depth of the $g$-network, respectively.
Both the $f$- and $g$-networks have a linear final layer and ReLU activation functions for all previous layers.
The hidden state $\h_0$ is initialized to zero.

To define the parameter search space, we define ranges for each parameter. For all tasks, we use $n \in \{ 1, 2, \ldots, 128 \}$, $\fwidth \in \{1, 2, \ldots, 256\}$, $\fdepth  \in \{ 1, 2, 3 \}$, $\gwidth \in \{1, 2, \ldots, 256\}$ and $\gdepth \in \{ 1, 2, 3 \}$, so the total search space consists of $128 \times 256 \times 3 \times 256 \times 3 = 75,497,472$ hyperparameter vectors $\hyperpars_i$.
 We order this search space by imposing a strict ordering on the importance of minimizing each hyperparameter 
 -- lower $\gdepth$ is strictly more important than lower $\fdepth$, which is strictly more important than lower $\nhidden$, which is strictly more important than lower $\gwidth$, which is strictly more important than lower $\fwidth$.
 We aim to find the hyperparameter vector (integer 5-tuple) $p_i$ in the search space which has lowest rank $i$ under this ordering. 

We search the space in the following simple manner. We first start at index $i = 65,536$, which corresponds to parameters $(1, 1, 2, 1, 1)$.
For each parameter tuple, we train networks using 5 different seeds. We use the loss function $\ell(x, y) = \frac{1}{2}\log[1 + (x-y)^2]$, finding that it led to more stable training than using vanilla MSE loss. We train for either 10,000 or 20,000 steps, depending on the task, using the Adam optimizer, a learning rate of $10^{-3}$, and batch size 4096. The test accuracy is evaluated with a batch of 65536 samples. If no networks achieve 100\% test accuracy (on any test batch), we increase $i$ by $2^{1/4}$. We proceed in this manner until we find a network where one of the seeds achieves perfect test accuracy or until the full range is exhausted. If we find a working network on this upwards sweep, we then perform a binary search using the interval halving method, starting from the successful $i$, to find the lowest $i$ where at least one seed achieves perfect test accuracy.

\subsection{Auto-simplification}

After finding a minimal neural network architecture that can solve a task, the resulting neural network weights typically seem random and un-interpretable. This is because there exist symmetry transformations of the weights that leave the overall input-output behavior of the neural network unchanged. The random initialization of the network has therefore caused random symmetry transformations to be applied to the weights.
In other words, the learned network belongs to an equivalence class of neural networks with identical behavior and performance, corresponding to a submanifold of the parameter space.
We exploit these symmetry transformations to simplify the neural network into a \textit{normal form}, which in a sense is the simplest member of its equivalence class.
Conversion of objects into a normal/standard form is a common concept in mathematics and physics (for example, conjunctive normal form, wavefunction normalization, reduced row echelon form, and gauge fixing).

\def\A{{\bf A}}
\def\C{{\bf C}}
\def\U{{\bf U}}
\def\V{{\bf V}}
\def\W{{\bf W}}
\def\b{{\bf b}}
\def\c{{\bf c}}
\def\h{{\bf h}}

Two of our simplification strategies below
exploit a symmetry of the RNN hidden state space $\h$. 
We can always write the MLP $g$ in the form 
$g(\h) = G(\U\h + \c)$ for some function $G$.
So if $f$ is affine, \ie, of the form
$f(\h,\x)= \W\h + \V\x + \b$, 
then the symmetry transformation

$\W'\equiv\A\W\A^{-1}$, $\V'=\A\V$, $\U' = \U\A^{-1}$, $\h'\equiv\A\h$, $\b'=\A\b$
keeps the RNN in the same form:
%defined by equations~\ref{RNNeq1}-\ref{RNNeq2}:
\beqa{hSymmetryEq}
\h_i'&=&\A\h_i =\A\W\A^{-1}\A\h_{i-1} + \A\V\x_i + \A\b\nonumber\\
&=&\W'^{-1}\h'_{i-1} + \V'\x_i + \b',\\
\y_i&=&G(\U\h_i + \c) = G(\U\A^{-1}\h'_i + \c) \nonumber\\
&=&G(\U'\h'_i + \c).
\eeqa

We think of neural networks as nails, which can be hit by various auto-normalization hammers. Each hammer is an algorithm that applies transformations to the weights to remove degrees of freedom caused by extra symmetries or cleans the neural network up in some other way. In this section, we describe five normalizers we use to simplify our trained networks, termed ``Whitening", ``Jordan normal form", ``Toeplitz", ``De-bias", and ``Quantization". For every neural network, we always apply this sequence of normalizers in that specific order, for consistency. We describe them below and provide additional details about them in the Appendix \ref{sec:hammer_details}.
\begin{enumerate}

\item {\bf Whitening:} Just as we normalize input data to use for training neural networks, we would like activations in the hidden state space $\mathbf{h}_i$ to be normalized. To ensure normalization in all directions, we feed the training dataset into the RNN, collect all the hidden states, compute the uncentered covariance matrix $\C$, and then apply a whitening transform $\h\mapsto\C^{-1/2}\h$ to the hidden state space so that its new covariance becomes the identity matrix. This operation exists purely to provide better numerical stability to the next step.

\item {\bf Jordan normal form:} When the function $g$ is affine, we can apply the aforementioned symmetry transformation to try to diagonalize $\W$, 
so that none of the hidden state dimensions interact with one another. 
Unfortunately, not all matrices $\W$ can be diagonalized, so we use a generalized alternative: the Jordan normal form, which allows elements of the superdiagonal to be either zero or one. To eliminate complex numbers, we also apply $2\times 2$ unitary transformations to eigenvectors corresponding to conjugate pairs of complex eigenvalues afterward.
The aforementioned whitening is now ruined, but it helped make the Jordan normal form calculation more numerically stable.

\item {\bf Toeplitz:} Once $\W$ is in Jordan normal form, we divide it up into Jordan blocks and apply upper-triangular Toeplitz transformations to the dimensions belonging to each Jordan block. There is now an additional symmetry, corresponding to multiplying each Jordan block by an
upper triangular Toeplitz matrix, and we exploit the Toeplitz matrix that maximally simplifies the aforementioned $\V$-matrix.

\item {\bf De-bias:} Sometimes $\W$ is not full rank, and $\b$ has a component in the direction of the nullspace. In this case, the component can be removed, and the bias $\c$ can be adjusted to compensate.

\item {\bf Quantization:}
After applying all the previous normalizers, many of the weights may have become close to integers, but not exactly due to machine precision and training imperfections. Sometimes, depending on the task, all of the weights can become integers. We therefore round any weights that are within $\epsilon\equiv 0.01$ of an integer to that integer.% \todo{(from Chloe) just curious, what percentage of the hammered weights usually end up as integers?}

\end{enumerate}

\subsection{Boolean and integer autoencoders}

\begin{figure}
    \centering
    \includegraphics[width=1\linewidth]{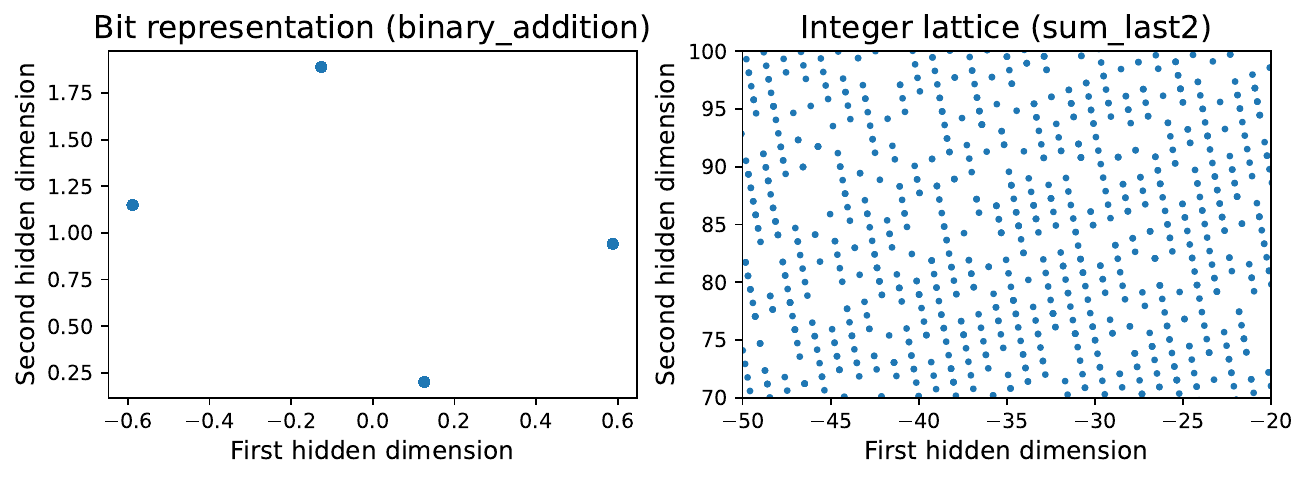}
    \vskip-4mm
    \caption{These hidden structures can be turned into discrete representations. 
Left: the hidden states for the bitstring addition task are seen to form four clusters, corresponding to 2 bits: the output bit and the carry bit.
Right: the hidden states for the Sum\_Last2 task are seen to form clusters on a 2D lattice corresponding to two integers.
}
\label{fig:bit_integer_illustration}
\end{figure}

As mentioned, our goal is to convert a trained recurrent neural network (RNN) into a maximally simple (Python) program that produces equivalent input-output behavior. This means that if the RNN has 100\% accuracy for a given dataset, so should the program, with the added benefit of being more interpretable, precise, and verifiable.

Once trained/written, the greatest difference between a neural network and a program implementing the same finite state machine is that the former is fuzzy and continuous, while the latter is precise and discrete. To convert a neural network to a program, some discretization (``defuzzification") process is needed to extract precise information from seemingly noisy representations. 
Fortunately, mechanistic interpretability research has shown that neural networks tend to learn meaningful, structured knowledge representations for algorithmic tasks~\cite{liu2022towards,nanda2023progress}. Previous interpretability efforts typically involved case-by-case manual inspection, and only gained algorithmic understanding at the level of pseudocode at best. We tackle this more ambitious question: can we create an automated method that distills the learned representation and associated algorithms into an equivalent (Python) program?

Since the tasks in our benchmark involve bits and integers, which are already discrete, the only non-discrete parts in a recurrent neural network are its hidden representations. Here we show two cases when hidden states can be discretized: they are  (1) a bit representation or (2) a (typically incomplete) integer lattice. Generalizing to the mixed case of bits and integers is straightforward.
\Fig{fig:bit_integer_illustration} shows all hidden state activation vectors $\h_i$ for all steps with all training examples for two of our tasks. 
The left panel shows that the $10^4$ points $\h_i$ form $2^2 = 4$ tight clusters, which we interpret as representing 2 bits. 
The right panel reveals that the points $\h_i$ form an incomplete 2D lattice that we interpret as secretly representing a pair of integers.

{\bf Bit representations}

The hidden states for the 2 bits in \Fig{fig:bit_integer_illustration} are seen to form a parallelogram.
More generally, we find that hidden states encode $b$ bits as $2^b$ clusters, which in some cases form $b$-dimensional parallelograms and in other cases look more random. Our algorithm tries all $(2^b)!$ possible assignments of the $2^b$ clusters to bitstrings of length $b$ and selects the assignment that minimizes the length of the resulting Python program.

{\bf Integer lattice} 

As seen in \Fig{fig:bit_integer_illustration}, the learned representation of an integer lattice tends to be both 
non-square (deformed by a random affine transformation) and sparse (since not all integer tuplets occur during training).
We thus face the following problem: given (possibly sparse) samples of points $\h_i$ from an $n$-dimensional lattice, how can we reconstruct the integer lattice in the sense that we figure out which integer tuple each lattice point represents? 
We call the solution an {\it integer autoencoder} since it compresses any point set into a set of integer tuples from which the original points can be at least approximately recovered as $\h_i = \A\k_i+\b$, where $\A$ is a matrix and $\b$ is a vector that defines the affine transformation and a set of integer vectors $\k_i$.

In the Appendix \ref{app:gcd}, we present a solution that we call the \textit{GCD lattice finder}.
For the special case $n=1$, its core idea is to compute the greatest common denominator of pairwise separations: for example, for the points 
$\{1.7,3.2,6.2,7.7...\}$, all point separations are divisible by $A=1.5$, from which one infers that $b=0.2$ and the lattice can be rewritten as $1.5\times $\{1,2,4,5\}$ + 0.2$.
For multidimensional lattices, our algorithm uses the GCD of ratios of generalized cell volumes to infer the directions and lengths of the lattice vectors that form the columns of $\A$. 

For the special case where the MLP defining the function $f$ is affine or can be accurately approximated as affine, we use a simpler method we term the 
\textit{Linear lattice finder}, also described in Appendix \ref{app:linear-lattice-finder}.
Here the idea is to exploit that the lattice is simply an affine transformation of a regular integer lattice (the input data), so we can simply ``read off" the desired lattice basis vectors from this affine transformation.

{\bf Symbolic regression}

Once the hidden states $\h_i$ have been successfully mapped 
to Boolean or integer tuples as described above, the functions $f$ and $g$ that specify the
learned RNN can be re-expressed as lookup tables, showing 
their Boolean/integer output tuple for each Boolean/integer input tuple. All that remains is now {\it symbolic regression}, \ie, discovering the simplest possible 
symbolic formulae that define $f$ and $g$.

{\bf Boolean regression:} In the case where a function maps bits to a bit, our algorithm determines the following set of correct Boolean formulae and then returns the shortest one.
The first candidate formula is the function written in disjunctive normal form, which is always possible.
If the Boolean function is {\it symmetric}, \ie, invariant under all permutations of its arguments, then we also write it as an integer function of its bit sum.

{\bf Integer regression:}
In the case when a function maps integers to an integer,
we try the following two methods:
\begin{enumerate}
    \item If the function is linear, then we perform simple linear regression, round the resulting coefficients to integers, and simplify, \eg,  multiplications by 0 and 1.
    \item Otherwise, we use the brute-force symbolic solver from {\it AI Feynman}~\citep{udrescu2020ai2}, including the 6 unary operations $\{>,<,\sim,H,D,A\}$ and 4 binary operations $\{+,-,*,\%\}$ whose meanings are explained in Appendix~\ref{app:symbolic-regression}, then convert the simplest discovered formula into Python format.
\end{enumerate}
Once symbolic formulas have been separately discovered for each component of the vector-valued functions $f$ and $g$, we insert them into a template Python program that implements the basic loop over inputs that are inherent in an RNN. 
We present examples of our auto-generated programs in Figures \ref{fig:bit_addition_program} and \ref{fig:hammer_comparison} and in Appendix \ref{app:generated-programs}.

\section{Results}

We will now test the program synthesis abilities of our MIPS algorithm on a benchmark of algorithmic tasks specified by numerical examples. For comparison, we try the same benchmark on GPT-4 Turbo, which is currently (as of January 2024) described by OpenAI as their latest generation model, with a
128k context window and more capable than the original GPT-4.

\subsection{Benchmark}

Our benchmark consists of the 62 algorithmic tasks listed in Table~\ref{tab:tasks}.
They each map one or two integer lists of length 10 or 20 into a new integer list. 
We refer to integers whose range is limited to $\{0,1\}$ as bits.
We generated this task list manually, attempting to produce a collection of diverse tasks that would in principle be solvable by an RNN. 
We also focused on tasks whose known algorithms involved majority, minimum, maximum, and absolute value functions because we believed they would be more easily learnable than other nonlinear algorithms due to our choice of the ReLU activation for our RNNs.
The benchmark training data and project code is available at \url{https://github.com/ejmichaud/neural-verification}.
The tasks are described in Table~\ref{tab:tasks}, with additional details in Appendix \ref{extra_task_explanation}.

Since the focus of our paper is not on whether RNNs can learn algorithms, but on whether learned algorithms can be auto-extracted into Python, we discarded from our benchmark any generated tasks on which our RNN-training failed to achieve 100\% accuracy.

Our benchmark can never show that MIPS outperforms any large language model (LLM). Because LLMs are typically trained on GitHub, many LLMs can produce Python code for complicated programming tasks that fall outside of the class we study. Instead, the question that our MIPS-LLM comparison addresses is whether MIPS complements LLMs by being able to solve some tasks where an LLM fails.

\subsection{Evaluation}

\begin{figure}[htbp]
\resizebox{\columnwidth}{!}{% Resize to fit within a single column
\centering
\begin{tikzpicture}[
    auto,
    node distance=0.5cm and 0.25cm, % vertical and horizontal distance between nodes
    >=Latex, % arrow tip style
    % Style for different types of nodes
    start/.style={draw, rectangle, fill=green!30, text width=5cm, text centered, minimum height=2em},
    io/.style={draw, rectangle, rounded corners, fill=blue!30, text width=3cm, text centered, minimum height=3em},
    process/.style={draw, rectangle, fill=orange!30, text width=5cm, text centered, minimum height=3em},
    decision/.style={draw, diamond, aspect=2, fill=yellow!30, text width=5cm, text badly centered, inner sep=0pt},
    arrow/.style={->, thick},
    dottedline/.style={draw, thick, dotted, -Latex}]

% Nodes
\node (start) [start, xshift=-5cm] {Conversation Start};
\node (user1) [io, below=of start] {User: "Each row in the table below contains two lists...give me a formula for ..."};
\node (gpt1) [process, below=of user1] {GPT: [Response]};
\node (user2) [io, below=of gpt1] {User: "Please write a Python program to ..."};
\node (gpt2) [process, below=of user2] {GPT: [Response]};
\node (codeBlock) [process, below=of gpt2, ] {Extracted Code Block};
\node (decision) [decision, below=of codeBlock] {Success or Failure?};
\node (success) [start, below left=of decision, xshift=-0.02cm] {Success};
\node (failure) [start, below right=of decision, xshift=0.02cm] {Failure};

% Arrows
\draw [arrow] (start) -- (user1);
\draw [arrow] (user1) -- (gpt1);
\draw [arrow] (gpt1) -- (user2);
\draw [arrow] (user2) -- (gpt2);
\draw [dottedline] (gpt2) -- (codeBlock);
\draw [dottedline] (codeBlock) -- (decision);
\draw [arrow] (decision) -- (success);
\draw [arrow] (decision) -- (failure);
\end{tikzpicture}}

\caption{We compare MIPS against program synthesis with the large language model GPT-4 Turbo, prompted with a ``chain-of-thought" approach. It begins with the user providing a task, followed by the model's response, and culminates in assessing the success or failure of the generated Python code based on its accuracy in processing the provided lists.}
\label{fig:code_generation_process}
\end{figure}
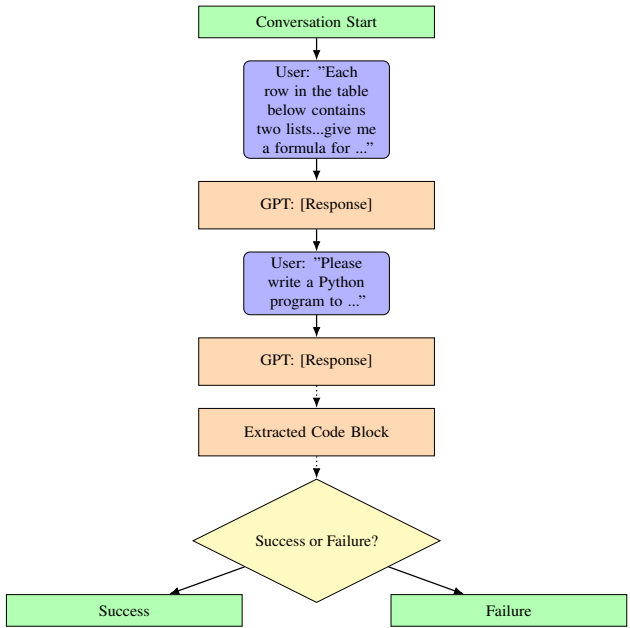

For both our method and GPT-4 Turbo, a task is considered solved if and only if a Python program is produced that solves the task with 100\% accuracy. GPT-4 Turbo is prompted using the ``chain-of-thought'' approach described below and illustrated in~\Cref{fig:code_generation_process}. 

For a given task, the LLM receives two lists of length 10 
%\todo{(from chloe) just to clarify, GPT4 sees less training examples than the automatic code extractor? I guess that's fine. idk} 
sourced from the respective RNN training set. The model is instructed to generate a formula that transforms the elements of list ``x" (features)  into the elements of list `y' (labels). 
Subsequently, the model is instructed to translate this formula into Python code. The model is specifically asked to use elements of the aforementioned lists as a test case and print ``Success" or ``Failure" if the generated function achieves full accuracy on the test case. An external program extracts a fenced markdown codeblock from the output, which is saved to a separate file and executed to determine if it successfully completes the task. To improve the chance of success, this GPT-4 Turbo prompting process is repeated three times, requiring only at least one of them to succeed. We run GPT using default temperature settings.

\begin{table*}[t]
\caption{Benchmark results. For tasks with note ``see text'', please refer to Appendix \ref{extra_task_explanation}}
\label{tab:tasks}
\vskip 0.05in
\begin{center}
\begin{footnotesize}
%\begin{sc}
\renewcommand{\arraystretch}{0.95} % Reducing vertical spacing
\begin{tabular}{@{}p{1.1cm}p{1cm}p{1.1cm}llp{1.5cm}p{1.4cm}@{}}
\toprule
Task \# & Input Strings & Element Type & Task Description & Task Name & Solved by GPT-4? & Solved by MIPS? \\ 
\midrule             
        1 & 2 & bit & Binary addition of two bit strings & Binary\_Addition & 0 & 1 \\ 
        2 & 2 & int & Ternary addition of two digit strings & Base\_3\_Addition & 0 & 0 \\ 
        3 & 2 & int & Base 4 addition of two digit strings & Base\_4\_Addition & 0 & 0 \\ 
        4 & 2 & int & Base 5 addition of two digit strings & Base\_5\_Addition & 0 & 0 \\ 
        5 & 2 & int & Base 6 addition of two digit strings & Base\_6\_Addition & 1 & 0 \\ 
        6 & 2 & int & Base 7 addition of two digit strings & Base\_7\_Addition & 0 & 0 \\ 
        7 & 2 & bit & Bitwise XOR & Bitwise\_Xor & 1 & 1 \\ 
        8 & 2 & bit & Bitwise OR & Bitwise\_Or & 1 & 1 \\ 
        9 & 2 & bit & Bitwise AND & Bitwise\_And & 1 & 1 \\ 
        10 & 1 & bit & Bitwise NOT & Bitwise\_Not & 1 & 1 \\ 
        11 & 1 & bit & Parity of last 2 bits & Parity\_Last2 & 1 & 1 \\ 
        12 & 1 & bit & Parity of last 3 bits & Parity\_Last3 & 0 & 1 \\ 
        13 & 1 & bit & Parity of last 4 bits & Parity\_Last4 & 0 & 0 \\ 
        14 & 1 & bit & Parity of all bits seen so far & Parity\_All & 0 & 1 \\ 
        15 & 1 & bit & Parity of number of zeros seen so far & Parity\_Zeros & 0 & 1 \\ 
        16 & 1 & int & Cumulative number of even numbers & Evens\_Counter & 0 & 0 \\ 
        17 & 1 & int & Cumulative sum & Sum\_All & 1 & 1 \\ 
        18 & 1 & int & Sum of last 2 numbers & Sum\_Last2 & 0 & 1 \\ 
        19 & 1 & int & Sum of last 3 numbers & Sum\_Last3 & 0 & 1 \\ 
        20 & 1 & int & Sum of last 4 numbers & Sum\_Last4 & 1 & 1 \\ 
        21 & 1 & int & Sum of last 5 numbers & Sum\_Last5 & 1 & 1 \\ 
        22 & 1 & int & sum of last 6 numbers & Sum\_Last6 & 1 & 1 \\ 
        23 & 1 & int & Sum of last 7 numbers & Sum\_Last7 & 1 & 1 \\ 
        24 & 1 & int & Current number & Current\_Number & 1 & 1 \\ 
        25 & 1 & int & Number 1 step back & Prev1 & 1 & 1 \\ 
        26 & 1 & int & Number 2 steps back & Prev2 & 1 & 1 \\ 
        27 & 1 & int & Number 3 steps back & Prev3 & 1 & 1 \\ 
        28 & 1 & int & Number 4 steps back & Prev4 & 1 & 1 \\ 
        29 & 1 & int & Number 5 steps back & Prev5 & 1 & 1 \\ 
        30 & 1 & int & 1 if last two numbers are equal & Previous\_Equals\_Current & 0 & 1 \\ 
        31 & 1 & int & $\text{current} - \text{previous}$ & Diff\_Last2 & 0 & 1 \\ 
        32 & 1 & int & $\lvert 	\text{current} - \text{previous} \rvert $ & Abs\_Diff & 0 & 1 \\ 
        33 & 1 & int & $\lvert \text{current} \rvert $ & Abs\_Current & 1 & 1 \\ 
        34 & 1 & int & $\lvert \text{current} \rvert - \lvert \text{previous} \rvert$ & Diff\_Abs\_Values & 1 & 0 \\ 
        35 & 1 & int & Minimum of numbers seen so far & Min\_Seen & 1 & 0 \\ 
        36 & 1 & int & Maximum of integers seen so far & Max\_Seen & 1 & 0 \\ 
        37 & 1 & int & integer in 0-1 with  highest frequency & Majority\_0\_1 & 1 & 0 \\ 
        38 & 1 & int & Integer in 0-2 with highest frequency & Majority\_0\_2 & 0 & 0 \\ 
        39 & 1 & int & Integer in 0-3 with highest frequency & Majority\_0\_3 & 0 & 0 \\ 
        40 & 1 & int & 1 if even, otherwise 0 & Evens\_Detector & 1 & 0 \\ 
        41 & 1 & int & 1 if perfect square, otherwise 0 & Perfect\_Square\_Detector & 0 & 0 \\ 
        42 & 1 & bit & 1 if bit string seen so far is a palindrome & Bit\_Palindrome & 1 & 0 \\ 
        43 & 1 & bit & 1 if parentheses balanced so far, else 0 & Balanced\_Parenthesis & 0 & 0 \\ 
        44 & 1 & bit & Number of bits seen so far mod 2 & Parity\_Bits\_Mod2 & 1 & 0 \\ 
        45 & 1 & bit & 1 if last 3 bits alternate & Alternating\_Last3 & 0 & 0 \\ 
        46 & 1 & bit & 1 if last 4 bits alternate & Alternating\_Last4 & 1 & 0 \\ 
        47 & 1 & bit & bit shift to right (same as prev1) & Bit\_Shift\_Right & 1 & 1 \\ 
        48 & 2 & bit & Cumulative dot product of bits mod 2 & Bit\_Dot\_Prod\_Mod2 & 0 & 1 \\ 
        49 & 1 & bit & Binary division by 3 (see text) & Div\_3 & 1 & 0 \\ 
        50 & 1 & bit & Binary division by 5 (see text) & Div\_5 & 0 & 0 \\ 
        51 & 1 & bit & Binary division by 7 (see text) & Div\_7 & 0 & 0 \\ 
        52 & 1 & int & Cumulative addition modulo 3 & Add\_Mod\_3 & 1 & 1 \\ 
        53 & 1 & int & Cumulative addition modulo 4 & Add\_Mod\_4 & 0 & 0 \\ 
        54 & 1 & int & Cumulative addition modulo 5 & Add\_Mod\_5 & 0 & 0 \\ 
        55 & 1 & int & Cumulative addition modulo 6 & Add\_Mod\_6 & 0 & 0 \\ 
        56 & 1 & int & Cumulative addition modulo 7 & Add\_Mod\_7 & 0 & 0 \\ 
        57 & 1 & int & Cumulative addition modulo 8 & Add\_Mod\_8 & 0 & 0 \\ 
        58 & 1 & int & 1D dithering, 4-bit to 1-bit (see text) & Dithering & 1 & 0 \\ 
        59 & 1 & int & Newton's of - freebody (integer input) & Newton\_Freebody & 0 & 1 \\ 
        60 & 1 & int & Newton's law of gravity (see text) & Newton\_Gravity & 0 & 1 \\ 
        61 & 1 & int & Newton's law w. spring (see text) & Newton\_Spring & 0 & 1 \\ 
        62 & 2 & int & Newton's law w. magnetic field (see text) & Newton\_Magnetic & 0 & 0 \\ 

\midrule
~ & ~ & ~ &Total solved& ~ & 30 & 32\\ 
\bottomrule
\end{tabular}%
%\end{sc}
\end{footnotesize}
\end{center}
\vskip -0.1in
\end{table*}

\clearpage

\subsection{Performance}

As seen in Table~1, MIPS is highly complementary to GPT-4 Turbo: 
MIPS solves 32 of our tasks, including 13 that are not solved by ChatGPT-4 (which solves 30). 

The AutoML process of \Cref{sec:automl} discovers networks of varying task-dependent shape and size. \Cref{architecture_search_table} shows the parameters $\hyperpars$ discovered for each task. Across our 62 tasks, 16 tasks could be solved by a network with hidden dimension $\nhidden = 1$, and the largest $\nhidden$ required was 81. For many tasks, there was an interpretable meaning to the shape of the smallest network we discovered. For instance, on tasks where the output is the element occurring $k$ steps earlier in the list, we found $n = k + 1$, since the current element and the previous $k$ elements must be stored for later recall.

We found two main failure modes for MIPS:
\begin{enumerate}
\item Noise and non-linearity. The latent space is still close to being a finite state machine, but the non-linearity and/or noise present in an RNN is so dominant that the integer autoencoder fails, \eg, for {\it Diff\_Abs\_Values}. Humans can stare at the lookup table and regress the symbolic function with their brains, but since the lookup table is not perfect, \ie, has the wrong integer in a few examples, MIPS fails to symbolically regress the function. This can probably be mitigated by learning and generalizing from a training subset with a smaller dynamic range. 
\item Continuous computation. A key assumption of MIPS is that RNNs are finite-state machines. However, RNNs can also use continuous variables to represent information --- the {\it Majority\_0\_X} tasks fail for this reason. This can probably be mitigated by identifying and implementing floating-point variables.
\end{enumerate}

\definecolor{codegreen}{rgb}{0,0.6,0}
\definecolor{codegray}{rgb}{0.5,0.5,0.5}
\definecolor{codepurple}{rgb}{0.58,0,0.82}
\definecolor{backcolour}{rgb}{0.95,0.95,0.92}
\lstdefinestyle{mystyle}{
    backgroundcolor=\color{backcolour},   
    commentstyle=\color{codegreen},
    keywordstyle=\color{magenta},
    numberstyle=\tiny\color{codegray},
    stringstyle=\color{codepurple},
    basicstyle=\ttfamily\footnotesize,
    breakatwhitespace=false,         
    breaklines=true,                 
    captionpos=b,                    
    keepspaces=true,                 
    numbers=left,                    
    numbersep=5pt,                  
    showspaces=false,                
    showstringspaces=false,
    showtabs=false,                  
    tabsize=2
}
\lstset{style=mystyle}

\begin{figure}

\definecolor{codegreen}{rgb}{0,0.6,0}
\definecolor{codegray}{rgb}{0.5,0.5,0.5}
\definecolor{codepurple}{rgb}{0.58,0,0.82}
\definecolor{backcolour}{rgb}{0.95,0.95,0.92}

\lstdefinestyle{mystyle}{
    backgroundcolor=\color{backcolour},   
    commentstyle=\color{codegreen},
    keywordstyle=\color{magenta},
    numberstyle=\tiny\color{codegray},
    stringstyle=\color{codepurple},
    basicstyle=\ttfamily\footnotesize,
    breakatwhitespace=false,         
    breaklines=true,                 
    captionpos=b,                    
    keepspaces=true,                 
    numbers=left,                    
    numbersep=5pt,                  
    showspaces=false,                
    showstringspaces=false,
    showtabs=false,                  
    tabsize=2
}

\lstset{style=mystyle}
        \begin{lstlisting}[language=Python]

def f(s,t):
    a = 0;b = 0;
    ys = []
    for i in range(10):
        c = s[i]; d = t[i];
        next_a = b ^ c ^ d
        next_b = b+c+d>1
        a = next_a;b = next_b;
        y = a
        ys.append(y)
    return ys \end{lstlisting}
    %     \centering
    %     \caption{With normalizers.}
    %     \label{fig:better_rnn_sum_last5_numerical}
    % \end{subfigure}
    \caption{The generated program for the addition of two binary numbers represented as bit sequences. Note that MIPS rediscovers the ``ripple adder", where the variable $b$ above is the carry bit.}
    \label{fig:bit_addition_program}
\end{figure}

\Fig{fig:bit_addition_program} shows an example of a MIPS rediscovering the "ripple-carry adder" algorithm.
The normalizers significantly simplified some of the resulting programs, as illustrated in \fig{fig:hammer_comparison}, and sometimes made the difference between MIPS failing and succeeding.
We found that applying a small $L1$ weight regularization sometimes facilitated integer autoencoding by axis-aligning the lattice.

\begin{figure}
    % \begin{subfigure}
        \begin{lstlisting}[language=Python]
def f(s):
    a = 198;b = -11;c = -3;d = 483;e = 0;
    ys = []
    for i in range(20):
        x = s[i]
        next_a = -b+c+190
        next_b = b-c-d-e+x+480
        next_c = b-e+8
        next_d = -b+e-x+472
        next_e = a+b-e-187
        a = next_a;b = next_b;c = next_c;d = next_d;e = next_e;
        y = -d+483
        ys.append(y)
    return ys\end{lstlisting}
        % \centering
        % \caption{Without normalizers.}
        % \label{fig:worse_rnn_sum_last5_numerical}
    % \end{subfigure}
    % \begin{subfigure}
        \begin{lstlisting}[language=Python]
def f(s):
    a = 0;b = 0;c = 0;d = 0;e = 0;
    ys = []
    for i in range(20):
        x = s[i]
        next_a = +x
        next_b = a
        next_c = b
        next_d = c
        next_e = d
        a = next_a;b = next_b;c = next_c;d = next_d;e = next_e;
        y = a+b+c+d+e
        ys.append(y)
    return ys\end{lstlisting}
    %     \centering
    %     \caption{With normalizers.}
    %     \label{fig:better_rnn_sum_last5_numerical}
    % \end{subfigure}
    \caption{Comparison of code generated from an RNN trained on Sum\_Last5, with (top) and without (top) normalizers. The whitening normalizer provided numerical stability to the Jordan normal form normalizer, which itself simplified the recurrent portion of the program. The Toeplitz and de-biasing normalizers jointly sparsified the occurrences of $x$ in the program, and the number of terms required to compute $y$. The quantization normalizer enabled all variables to be represented as integers.}
    \label{fig:hammer_comparison}
\end{figure}

%* Are there any interesting patterns for where GPT-4 beats MIPS or vice versa?

% \usepackage{booktabs}
% \usepackage{graphicx}

\section{Conclusions}

% WHAT WE'VE DONE

We have presented MIPS, a novel method for program synthesis based on automated mechanistic interpretability of neural networks trained to perform the desired task, auto-distilling the learned algorithm into Python code.
Its essence is to first train a recurrent neural network to learn a clever finite state machine that performs the task, and then automatically figure out how this machine works.

\subsection{Findings}

We found MIPS highly complementary to LLM-based program synthesis with GPT-4 Turbo, with each approach solving many tasks that stumped the other. 
Whereas LLM-based methods have the advantage of drawing upon a vast corpus of human training data, MIPS has the advantage of discovering algorithms from scratch without human hints, with the potential to discover entirely new algorithms. As opposed to genetic programming approaches, MIPS leverages the power of deep learning by exploiting gradient information.

Program synthesis aside, our results shed further light on mechanistic interpretability, specifically on how neural networks represent bits and integers. We found that $n$ integers tend to get encoded linearly in $n$ dimensions, but generically in non-orthogonal directions with an additive offset. This is presumably because there are many more such messy encodings than simple ones, and the messiness can be easily (linearly) decoded. We saw that $n$ bits sometimes get encoded as an $n$-dimensional parallelogram, but not always ––– possibly because linear decodability is less helpful when the subsequent bit operations to be performed are nonlinear anyway.

\subsection{Outlook}

Our work is merely a modest first attempt at mechanistic-interpretability-based program synthesis, and there are many obvious generalizations worth trying in future work. For example:
\begin{enumerate}
\item Improvements in training and integer autoencoding (since many of our failed examples failed only just barely)
\item Generalization from RNNs to other architectures such as transformers
\item Generalization from bits and integers to more general extractable data types such as floating-point numbers and various discrete mathematical structures and knowledge representations
\item Scaling to tasks requiring much larger neural networks
\item Automated formal verification of synthesized programs (we perform such verification with Dafny 
in \Cref{app:dafnyverification} to show that our MIPS-learned ripple adder correctly adds {\it any} binary numbers, not merely those in the test set, but such manual work should ideally be fully automated)
\end{enumerate}

LLM-based coding co-pilots are already highly useful for program synthesis tasks based on verbal problem descriptions or auto-complete, and will only get better. MIPS instead tackles program synthesis based on test cases alone. 
This makes it analogous to {\it symbolic regression} \cite{udrescu2020ai2,cranmer2023interpretable}, which has already proven useful for various science and engineering applications \cite{cranmer2020discovering,ma2022evolving} where one wishes to approximate data relationships with symbolic formulae. 
The MIPS framework generalizes symbolic regression from feed-forward formulae to programs with loops, which are in principle Turing complete.
If this approach can be scaled up, it may enable promising opportunities for making machine-learned algorithms more interpretable, verifiable, and trustworthy.

\section*{Broader Impact}

Because machine-learned algorithms now outperform traditional human-discovered algorithms on many tasks, there are incentives to deploy them even without a full understanding of how they work and of whether they are biased, unsafe, or otherwise problematic. The aspirational broader impact motivating this paper is to help automate the process of making AI systems more transparent, robust, and trustworthy, with the ultimate goal of developing provably safe AI systems 
\cite{tegmark2023provably}.

\section*{Acknowledgements}

%{\bf [COMMENT OUT FOR ANONYMOUS SUBMISION!]}
We thank Wes Gurnee, James Liu, and Armaun Sanayei for helpful conversations and suggestions. This work is supported by Erik Otto, Jaan Tallinn, the Rothberg Family Fund for Cognitive Science, the NSF Graduate Research Fellowship (Grant No.
2141064), and IAIFI through NSF grant PHY-2019786.

\bibliography{refs}
\bibliographystyle{icml2024}

%* Zimings sharp-fuzzy intuition

\clearpage

\appendix

\onecolumn

%%%%%%%%%%%%%%%%%%%%%%%%%%%%%%%%%%%%%

\begin{table*}[t]
\caption{AutoML architecture search results. All networks achieved 100\% accuracy on at least one test batch.}
\label{architecture_search_table}
\vskip 0.15in
\begin{center}
\begin{small}
% \begin{sc}
\renewcommand{\arraystretch}{0.97} % Reducing vertical spacing
\begin{tabular}{llccccccc}
\toprule
Task \# & Task Name & $\nhidden$ & $\fwidth$ & $\fdepth$ & $\gwidth$ & $\gdepth$ & Train Loss & Test Loss \\
\midrule
% Include the lines generated by your Python script here
\input{tables/architecture_search_results}
% \midrule
\end{tabular}
% \end{sc}
\end{small}
\end{center}
\vskip -0.1in
\end{table*}

\section{Lattice finding using generalized greatest common divisor (GCD)} \label{app:gcd}

Our method often encounters cases where hidden states secretly form an affine transformation of an integer lattice. However, not all lattice points are observed in training samples, so our goal is to recover the hidden integer lattice from sparse observations.

\subsection{Problem formulation}
Suppose we have a set of lattice points in $\mathbb{R}^D$
spanned by $D$ independent basis vectors, $\mat{b}_i \,(i=1,2,\cdots,D)$. Each lattice point $j$ has the position
\begin{equation}\label{eq:lat}
    \mat{x}_j = \sum_{i=1}^D a_{ji} \mat{b}_i + \mat{c},
\end{equation}
where $\mat{c}$ is a global translation vector, and the coefficients $a_{ji}$ are integers

Our problem: given $N$ such data points $(\mat{x}_1,\mat{x}_2,\cdots,\mat{x}_N)$, how can we recover the integer coefficients $a_{ji}$ for each point data point as well as $\mat{b}_i$ and $\mat{c}$?

Note that even when the whole lattice is given, there are still degrees of freedom for the solution. For example, $\{\mat{c}\mapsto \mat{c}+\mat{b}_i, a_{ji}\mapsto a_{ji}-1\}$ remains a solution, and $\{\mat{b}_i\to \sum_{j=1}^D\Lambda_{ij}\mat{b}_j\}$ remains a solution if $\mat{\Lambda}$ is an integer matrix whose determinant is $\pm 1$. So our success criterion is: (1) $a_{ji}$ are integers; (2) the discovered bases and the true bases have the same determinant (the volume of a unit cell). Once a set of bases is found, we can simplify them by minimizing their total norms over valid transformations ($\mat{\Lambda}\in\mathbb{Z}^{D\times D}, {\rm det}(\mat{\Lambda})=\pm 1$).

\subsection{Regular GCD}

As a reminder, given a list of $n$ numbers $\{y_1, y_2, \cdots, y_n\}$, a common divisor $d$ is a number such that for all $i$, $\frac{y_i}{d}$ is an integer. All common divisors are the set $\{d|y_i/d\in\mathbb{Z}$, and the greatest common divisor (GCD) is the largest number in this set. Because
\begin{equation}
    {\rm GCD}(y_1,\cdots,y_n) = {\rm GCD}(y_1, {\rm GCD}(y_2, {\rm GCD}(y_3,...))),
\end{equation}
it without loss of generality suffices to consider the  case $n=2$. A common algorithm to compute GCD of two number is the so-called Euclidean algorithm. We start with two numbers $r_0$, $r_1$ and $r_0 > r_1$, which is step 0. For the $k^{\rm th}$ step, we perform division-with-remainder to find the quotient $q_k$ and the remainder $r_k$ so that
$r_{k-2} = q_k r_{k-1} + r_k$ with $|r_{k-1}| > |r_{k}|$~\footnote{We are considering a general case where $r_0$ and $r_1$ may be negative. Otherwise $r_k$ can always be positive numbers, hence no need to use the absolute function.}. The algorithm will eventually produce a zero remainder $r_N=0$, and the other non-zero remainder $r_{N-1}$ is the greatest common divisor. 
For example, ${\rm GCD}(55,45)=5$, because
\begin{equation}
    \begin{aligned}
       55 & = 1 \times 45 + 10,\\
       45 & = 4 \times 10 + 5,\\
       10 & = 2 \times 5 + 0.
    \end{aligned}
\end{equation}

\subsection{Generalized GCD in $D$ dimensions}

Given a list of $n$ vectors $\{\mat{y}_1,\mat{y}_2,\cdots,\mat{y}_n\}$ where $\mat{y}_i\in\mathbb{R}^D$, and assuming that these vectors are in the lattice described by Eq.~(\ref{eq:lat}), we can without loss of generality set $\mat{c}=0$, since we can always redefine the origin. In $D$ dimensions, the primitive of interest is the $D$-dimensional parallelogram: a line segment for $D=1$ (one basis vector), a parallelogram for $D=2$ (two basis vectors), 
parallelepiped for $D=3$ (three basis vectors), \etc

One can construct a $D$-dimensional parallelogram by constructing its basis vectors as a linear integer combination of $\mat{y}_j$, i.e.,

\begin{equation}
    \mat{q}_i = \sum_{j=1}^n m_{ij}\mat{y}_j, m_{ij}\in\mathbb{Z}, i=1,2,\cdots, D.
\end{equation}
The goal of $D$-dimensional GCD is to find a ``minimal'' parallelogram, such that its volume (which is ${\rm det}(\mat{q}_1,\mat{q}_2,\cdots,\mat{q}_D)$) is GCD of volumes of other possible parallelograms. Once the minimal parallelogram is found~\footnote{There could be many minimal parallelograms, but finding one is sufficient.}, we can also determine $\mat{b}_i$ in Eq.~(\ref{eq:lat}), since $\mat{b}_i$ are exactly $\mat{q}_i$! To find the minimal parallelogram, we need two steps: (1) figure out the unit volume; (2) figure out $\mat{q}_i (i=1,2,\cdots)$ whose volume is the unit volume.

{\bf Step 1: Compute unit volume $V_0$.} We first define \textit{representative} parallelograms as one where all $i=1,2,\cdots,D$, $\mat{m}_i\equiv (m_{i1},m_{i2},\cdots,m_{iD})$ are one-hot vectors, \ie, with only one element being 1 and 0 otherwise. It is easy to show that the volume of any parallelogram is a linear integer combination of 
volumes of representative parallelograms, so WLOG we can focus on representative parallelograms. We compute the volumes of all representative parallelograms, which gives a volume array. Since volumes are just scalars, we can get the unit volume $V_0$ by calling the regular GCD of the volume array.

{\bf Step 2: Find a minimal parallelogram (whose volume is the unit volume computed in step 1).} Recall that in regular GCD, we are dealing with two numbers (scalars). To leverage this in the vector case, we need to create scalars out of vectors, and make sure that the vectors share the same linear structure as the scalars so that we can extend division-and-remainder to vectors. A natural scalar is volume. Now consider two parallelograms P1 and P2, which share $D-1$ basis vectors $(\mat{y}_3,\dots,\mat{y}_{D+1})$, but last basis vector is different: $\mat{y}_1$ for P1 and $\mat{y}_2$ for P2. Denote their volume as $V_1$ and $V_2$:
\begin{equation}
    \begin{aligned}
        V_1 &= {\rm det}(\mat{y}_1,\mat{y}_3,\mat{y}_4,\dots,\mat{y}_D) \\
        V_2 &= {\rm det}(\mat{y}_2,\mat{y}_3,\mat{y}_4,\dots,\mat{y}_D) \\
    \end{aligned}
\end{equation}
Since
\begin{equation}
    aV_1 + bV_2 = {\rm det}(a\mat{y}_1 + b\mat{y}_2,\mat{y}_3,\mat{y}_4,\dots,\mat{y}_D),
\end{equation}
which shows that $(V_1, V_2)$ and $(\mat{y}_1,\mat{y}_2)$ share the same linear structure. We can simply apply division-and-remainder to $V_1$ and $V_2$ as in regular GCD:
\begin{equation}
    V_1', V_2' = {\rm GCD}(V_1, V_2),
\end{equation}
whose quotients in all iterations are saved and transferred to $\mat{y}_1$ and $\mat{y}_2$:
\begin{equation}
    \mat{y}_1',\mat{y}_2' = {\rm GCD\_with\_predefined\_quotients}(\mat{y}_1,\mat{y}_2).
\end{equation}
If $V_1 = V_0$ (which is the condition for minimal parallelogram), the algorithm terminates and returns $(\mat{y}_1',\mat{y}_3,\mat{y}_4,\cdots,\mat{y}_D)$. If $V_1>V_0$, we need to repeat step 2 with the new vector list $\{\mat{y}_1',\mat{y}_3,\cdots,\mat{y}_N\}$.

{\bf Why can we remove $\mat{y}_2'$ for next iteration?} Note that although eventually $V_1'>0$ and $V_2'=0$, typically $\mat{y}_2\neq 0$. However, since 
\begin{equation}
  0 = V_2' = {\rm det}(\mat{y}_2', \mat{y}_3,\mat{y}_4,\cdots,\mat{y}_D),
\end{equation}
this means $\mat{y}_2'$ is a linear combination of $(\mat{y}_3,\cdots,\mat{y}_D)$, hence can be removed from the vector list.

\begin{figure}
    \centering
    \includegraphics[width=0.8\linewidth]{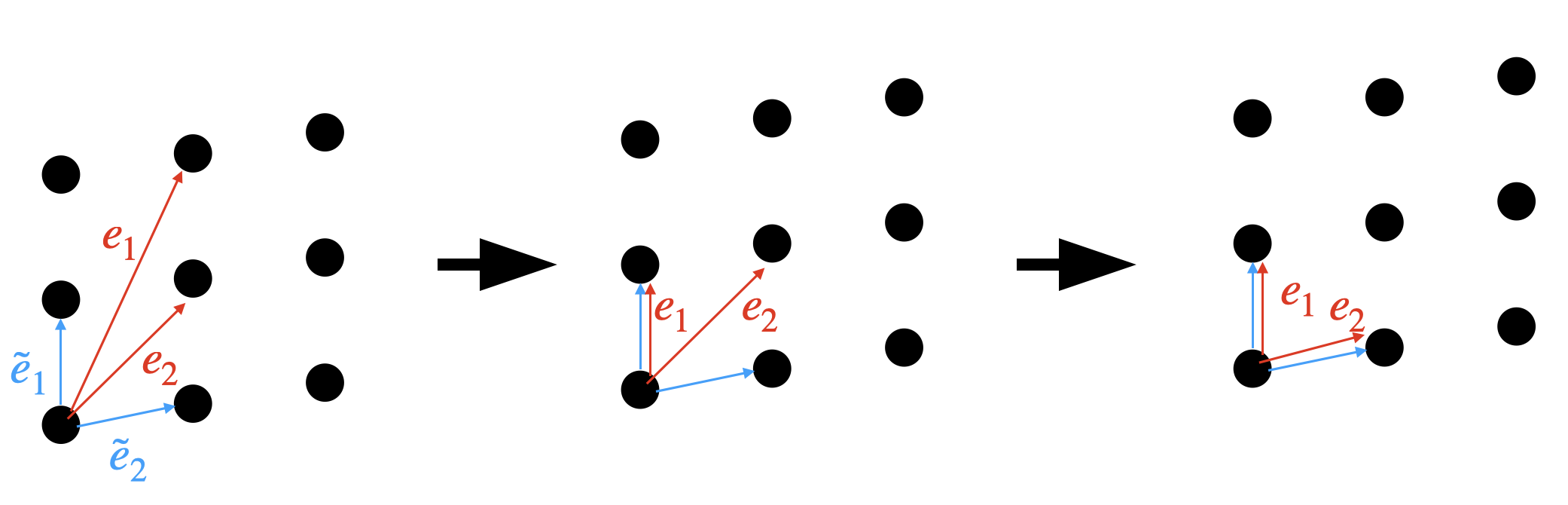}
    \caption{Both red and blue basis form a minimal parallelogram (in terms of cell volume), but one can further simplify red to blue by linear combination (simplicity in the sense of small $\ell_2$ norm).}
    \label{fig:lattice-simplification}
\end{figure}

{\bf Step 3: Simplification of basis vectors.} We want to further simplify basis vectors. For example, the basis vectors obtained in step 2 may have large norms. For example $D=2$, the standard integer lattice has $\mat{b}_1 = (1,0)$ and $\mat{b}_2=(0,1)$, but they are infinitely many possibilities after step 2, as long as $pt-sq=\pm 1$ for $\mat{b}_1 = (p,q)$ and $\mat{b}_2=(s,t)$, e.g., $\mat{b}_1 = (3,5)$ and $\mat{b}_2 = (4,7)$. 

To minimize $\ell_2$ norms, we choose a basis and project-and-subtract for other bases. Note that: (1) again we are only allowed to subtract integer times of the chosen basis; (2) the volume of the parallelogram does not change since the project-and-subtract matrix has determinant 1 (suppose $\mat{b}_i (i=2,3,\cdots,D)$ are projected to $\mat{b}_1$ and subtracted by multiples of $\mat{b}_1$. $p_{*}$ represents projection integers):

\begin{equation}
    \begin{pmatrix}
        1 & p_{2\to 1} & p_{3\to 1} & \cdots & p_{D\to 1} \\
        0 & 1 & 0 & \cdots & 0 \\
        0 & 0 & 1 & \cdots & 0 \\
        \vdots & \vdots & \vdots & & \vdots \\
        0 & 0 & 0 & \cdots & 1
    \end{pmatrix}
\end{equation}

We do this iteratively, until no norm can become shorter via project-and-subtract. Please see Figure~\ref{fig:lattice-simplification} for an illustration of how simplification works for a 2D example.

{\bf Computation overhead} is actually surprisingly small. In typical cases, we only need to call $O(1)$ times of GCD.

{\bf Dealing with noise} Usually the integer lattice in the hidden space is not perfect, i.e., vulnerable to noise. How do we extract integer lattices in a robust way in the presence of noise? Note that the terminating condition for the GCD algorithm is when the remainder is exactly zero - we relax this condition to that the absolute value of the remainder to be smaller than a threshold $\epsilon_{\rm gcd}$. Another issue regarding noise is that noise can be accumulated in the GCD iterations, so we hope that GCD can converge in a few steps. To achieve this, we select hidden states in a small region with data fraction $p\%$ of the whole data. Since both $\epsilon_{\rm gcd}$ and $p$ depends on data and neural network training which we do not know a priori, we choose to grid sweep $\epsilon_{\rm gcd}\in [10^{-3},1]$ and $p\in (0.1,100)$; for each $(\epsilon_{\rm gcd},p)$, we obtain an integer lattice and compute its description length. We select the $(\epsilon_{\rm gcd},p)$ which gives the lattice with the smallest description length. The description length includes two parts: integer descriptions of hidden states ${\rm log}(1+|Z|^2)$, and residual of reconstruction ${\rm log}(1+(\frac{AZ+b-X}{\epsilon_{\rm dl}})^2)$ with $\epsilon_{\rm dl}=10^{-4}$.

\section{Linear lattice finder}\label{app:linear-lattice-finder}

Although our RNN can represent general nonlinear functions, in the special case when the RNN actually performs linear functions, program synthesis can be much easier. So if the hidden MLP is linear, we would expect the hidden states to be an integer lattice, because inputs are integer lattices and the mappings are linear. Effectively the hidden MLP works as a linear function: $\h^{(t)}=W_h\h^{(t-1)}+W_i\x^{(t)}$ (we neglected the bias term since it is not relevant to finding basis vectors of a lattice).

Suppose we have input series $\x^{(1)}, \x^{(2)}, ..., \x^{(t)}$, then $\h^{(t)}$ is 
\begin{equation}
    \h^{(t)} = \sum_{j=1}^t W_h^{t-j}W_i\x_{j},
\end{equation}
Since $\x_j$ themselves are integer lattices, we could then interpret the following as basis vectors:
\begin{equation}
    W_h^{t-j}W_i, j = 1,2,\cdots, t,
\end{equation}
which are not necessarily independent. For example, for the task of summing up the last two numbers, $W_hW_i$ and $W_i$ are non-zero vectors and are independent, while others $W_h^nW_i\approx 0, n\geq 2$. Then $W_hW_i$ and $W_i$ are the two basis vectors for the lattice. In general, we measure the norm of all the candidate basis vectors, and select the first $k$ vectors with highest norms, which are exactly basis vectors of the hidden lattice.

\section{Symbolic regression}\label{app:symbolic-regression}
The formulation of symbolic regression is that one has data pair $(\x_i, y_i), i=1,2,\dots,N$ with $N$ data samples. The goal is to find a symbolic formula $f$ such that $y_i=f(\x_i)$. A function is expressed in reverse polish notation (RPN), for example, $|a|-c$ is expressed as \texttt{aAc-} where $A$ stands for the absolute value function. We have three types of variables:
\begin{itemize}
    \item type-0 operator. We include input variables and constants.
    \item type-1 operator (takes in one type-0 to produce one type-0). We include operations $\{>,<,\sim,H,D,A\}$. $>$ means $+1$; $<$ means $-1$; $\sim$ means negating the number; $D$ is dirac delta which outputs 1 only when taking in 0; $A$ is the absolute value function;
    \item type-2 operator (takes in two type-0 to produce on type-0). We include operations $\{+,*,-,\%\}$. $+$ means addition of two numbers; $*$ means multiplication of two numbers; $-$ means subtraction of two numbers; $\%$ is the remainder of one number module the other. 
\end{itemize}
There are only certain types of templates (a string of numbers consisting of 0,1,2) that are syntactically correct. For example, $002$ is correct while $02$ is incorrect. We iterate over all the templates not longer than 6 symbols, and for each template, we try all the variable combinations. Each variable combination corresponds to a symbolic equation $f$, for which we can check whether $f(\x_i)=y_i$ for 100 data points. If success, we terminate the brute force program and return the successful formula. If brute force search does not find any correct symbolic formula within compute budget, we will simply return the formula $a$, to make sure that the program can still be synthesized but simply fail to make correct predictions.

\section{Neural Network Normalization Algorithms} \label{sec:hammer_details}
It is well known that neural networks exhibit a large amount of symmetry. That is, there are many transformations that can be applied to networks without affecting the map $y = f(x)$ that they compute. A classic example is to permute the neurons within layers.

In this section, we describe a suite of normalizers that we use to transform our networks into a standard form, such that the algorithms that they learn are easier to interpret. We call our five normalizers ``Whitening'', ``Jordan normal form (JNF)'', ``Toeplitz'', ``De-bias'', and ``Quantization''.

The main symmetry which we focus on is a linear transformation of the hidden space $\h \mapsto \A\h$, which requires the following changes to $f$ and $g$:
\begin{align*}
    f(\h, \x) = \W\h + \V\x + \b \implies& f(\h, \x) = \A\W\A^{-1}\h + \A\V\x + \A\b \\
    g(\h) = G(\U\h + \c) \implies& f(\h) = G(\U\A^{-1}\h + \c)
\end{align*}
and is implemented by changing the weights:
\begin{align*}
    \W \implies& \A\W\A^{-1} \\
    \V \implies& \A\V \\
    \b \implies& \A\b \\
    \U \implies& \U\A^{-1}
\end{align*}

\def\S{\bf S}
\def\I{\bf I}

For this symmetry, we can apply an arbitrary invertible similarity transformation $\A$ to $\W$, which is the core idea underlying our normalizers, three of which have their own unique ways of constructing $\A$, as we describe in the sections below. Most importantly, one of our normalizers exploits $\A$ to convert the hidden-to-hidden transformation $\W$ into Jordan normal form, in the case where $f$ is linear. Recent work has shown that large recurrent networks with linear hidden-to-hidden transformations, such as state space models \citep{gu2023mamba} can perform just as well as transformer-based models in language modeling on a large scale. A main advantage of using linear hidden-to-hidden transformations is the possibility of expressing the hidden space in it's eigenbasis. This causes the hidden-to-hidden transformation to become diagonal, so that it can be computed more efficiently. In practice, modern state space models assume diagonality, and go further to assume the elements on the diagonal are real; they fix the architecture to be this way during training.

By doing this, we ignore the possibility of linear hidden-to-hidden transformations that cannot be transformed into a real diagonal matrix via diagonalization. Such examples include rotation matrices (whose eigenvalues may be complex), and shift matrices (whose eigenvalues are degenerate and whose eigenvectors are duplicated). A more general form than the diagonal form is the Jordan normal form, which consists of Jordan blocks along the diagonal, each of which has the form $\lambda \I + \S$ for an eigenvalue $\lambda$ and the shift matrix $\S$ with ones on the superdiagonal and zeros elsewhere. The diagonalization is a special case of Jordan normal form, and all matrices can be transformed to Jordan normal form. A simple transformation can also be applied to Jordan normal forms that contain pairs of complex generalized eigenvectors, to convert them into real matrices.

For nonlinear hidden-to-hidden transformations, we compute $\W$ as though the nonlinearities have been removed.

\subsection{Whitening Transformation}
Similar to normalizing the means and variances of a dataset before feeding it into a machine learning model, a good first preprocessing step is to normalize the distribution of hidden states. We therefore choose to apply a whitening transformation to the hidden space. To compute the transformation, we compute the covariance matrix of hidden activations across the dataset, and use the singular value decomposition (SVD) of this covariance matrix to find the closest transformation to the identity that will bring this covariance matrix to the identity. We ignore any directions with covariance less than $\epsilon=0.1$, which cause more instability when normalized. We then post-apply this transformation to the last linear layer of the hidden-to-hidden transformation and its biases, and pre-apply its inverse to the first layers of the hidden-to-hidden and hidden-to-output transformations. This leaves the net behavior of the network unchanged. Other transformations which we use in other normalizers operate in a similar manner, by post-applying and pre-applying a transformation and its inverse transformation to the first and last layers that interact with the hidden space.

\subsection{Jordan Normal Form Transformation}
Critically, the hidden-to-hidden transformations which we would like to convert into Jordan normal form are imperfect because they are learned. Eigenvectors belonging to each Jordan block must be identical, whereas this will only be approximately true of the learned transformation.

\def\T{\bf T}

The Jordan normal form of a matrix is unstable; consider a matrix
\begin{align*}
    \W = \begin{pmatrix} 0 & 1 \\ \delta & 0 \end{pmatrix}
\end{align*}
which, when $\delta \neq 0$, can be transformed into Jordan normal form by:
\begin{align}
    \begin{pmatrix} 0 & 1 \\ \delta & 0 \end{pmatrix} = \underbrace{\begin{pmatrix} 1 & 1 \\ \sqrt{\delta} & -\sqrt{\delta} \end{pmatrix}}_{\T} \begin{pmatrix} \sqrt{\delta} & 0 \\ 0 & -\sqrt{\delta} \end{pmatrix} \begin{pmatrix} 1 & 1 \\ \sqrt{\delta} & -\sqrt{\delta} \end{pmatrix}^{-1} \label{eq:worse_jnf}
\end{align}
but when $\delta = 0$, is transformed into Jordan normal form by:
\begin{align}
    \begin{pmatrix} 0 & 1 \\ 0 & 0 \end{pmatrix} = \underbrace{\begin{pmatrix} 1 & 0 \\ 0 & 1 \end{pmatrix}}_{\T} \begin{pmatrix} 0 & 1 \\ 0 & 0 \end{pmatrix} \begin{pmatrix} 1 & 0 \\ 0 & 1 \end{pmatrix}^{-1} \label{eq:better_jnf}
\end{align}
As we can see, all of the matrices in the decomposition are unstable near $\delta=0$, so the issue of error thresholding is not only numerical, but is mathematical in nature as well.

We would like to construct an algorithm which computes the Jordan normal form with an error threshold $|\delta| < \epsilon=0.7$ within which the algorithm will pick the transformation $\T$ from Equation (\ref{eq:better_jnf}) instead of from Equation (\ref{eq:worse_jnf}).

\def\X{{\bf X}}
\def\J{{\bf J}}
\def\F{{\bf F}}

Our algorithm first computes the eigenvalues $\lambda_i$, and then iteratively solves for the generalized eigenvectors which lie in $\text{ker}((\W-\lambda \I)^k)$ for increasing $k$. The approximation occurs whenever we compute the kernel (of unknown dimension) of a matrix $\X$; we take the SVD of $\X$ and treat any singular vectors as part of the nullspace if their singular values are lower than the threshold $\epsilon$, calling the result $\epsilon\text{-ker}(\X)$.

Spaces are always stored in the form of a rectangular matrix $\F$ of orthonormal vectors, and their dimension is always the width of the matrix. We build projections using $\text{proj}(\F) = \F\F^H$, where $\F^H$ denotes the conjugate transpose of $\F$. We compute kernels $\text{ker}(\X)$ of known dimension of matrices $\X$ by taking the SVD $\X=\V_1S\V_2^H$ and taking the last singular vectors in $\V_2^H$. We compute column spaces of projectors of known dimension by taking the top singular vectors of the SVD.

The steps in our algorithm are as follows:
\begin{enumerate}
    \item Solve for the eigenvalues $\lambda_i$ of $\W$, and check that eigenvalues that are within $\epsilon$ of each other form groups, ie. that $|\lambda_i-\lambda_j|\leq\epsilon$ and $|\lambda_j-\lambda_k|\leq\epsilon$ always implies $|\lambda_k-\lambda_i|\leq\epsilon$. Compute the mean eigenvalue for every group.
    \item Solve for the approximate kernels of $\W-\lambda \I$ for each mean eigenvalue $\lambda$. We will denote this operation by $\epsilon\text{-ker}(\W-\lambda \I)$. We represent these kernels by storing the singular vectors whose singular values are lower than $\epsilon$. Also, construct a ``corrected matrix'' of $\W-\lambda \I$ for every $\lambda$ by taking the SVD, discarding the low singular values, and multiplying the pruned decomposition back together again.
    \item Solve for successive spaces $\F_k$ of generalized eigenvectors at increasing depths $k$ along the set of Jordan chains with eigenvalue $\lambda$, for all $\lambda$. In other words, find chains of mutually orthogonal vectors which are mapped to zero after exactly $k$ applications of the map $\W-\lambda \I$. We first solve for $\F_0 = \text{ker}(\W-\lambda \I)$. Then for $k>0$, we first solve for $\J_k=\epsilon\text{-ker}((I-\text{proj}(\F_{k-1}))(\W-\lambda \I))$ and deduce the number of chains which reach depth $k$ from the dimension of $\J_k$, and then solve for $\F_k = \text{col}(\text{proj}(\J_k) - \text{proj}(\F_0))$.
    \item Perform a consistency check to verify that the dimensions of $\F_k$ always stay the same or decrease with $k$. Go through the spaces $\F_k$ in reverse order, and whenever the dimension of $\F_k$ decreases, figure out which direction(s) are not mapped to by applying $\W-\lambda \I$ to $\F_{k+1}$. Do this by building a projector $\J$ from mapping vectors representing $\F_{k+1}$ through $\W-\lambda \I$, and taking $\text{col}(\text{proj}(\F_k)-\J)$. Solve for the Jordan chain by repeatedly applying $\text{proj}(\F_i)(\W_i-\lambda \I)$ for $i$ starting from $k-1$ and going all the way down to zero.
    \item Concatenate all the Jordan chains together to form the transformation matrix $\T$.
\end{enumerate}

The transformation $\T$ consists of generalized eigenvectors which need not be completely real but may also include pairs of generalized eigenvectors that are complex conjugates of each other. Since we do not want the weights of our normalized network to be complex, we also apply a unitary transformation which changes any pair of complex generalized eigenvectors into a pair of real vectors, and the resulting block of $\W$ into a multiple of a rotation matrix. As an example, for a real 2 by 2 matrix $\W$ with complex eigenvectors, we have
\begin{align*}
\W &= \T\begin{pmatrix} a + bi & 0 \\ 0 & a - bi \end{pmatrix}\T^{-1} \\
&= \T\T'\begin{pmatrix} a & -b \\ b & a \end{pmatrix}(\T\T')^{-1}, \quad \T' = \frac{1}{\sqrt{2}}\begin{pmatrix} 1 & i \\ 1 & -i \end{pmatrix}
\end{align*}

\subsection{Toeplitz Transformation}
Once $\W$ is in Jordan normal form, each Jordan block is an upper triangular Toeplitz matrix. Upper-triangular Toeplitz matrices, including Jordan blocks, will always commute with each other, because they are all polynomials of the shift matrix (which has ones on the superdiagonal and zeros elsewhere,) and therefore these transformations will leave $\W$ unchanged, but will still affect $\V$. We split $\V$ up into parts operated on by each Jordan block, and use these Toeplitz transformations to reduce the most numerically stable columns of each block of $\V$ to one-hot vectors. The numerically stability of a column vector is determined by the absolute value of the bottom element of that column vector, since it's inverse will become the degenerate eigenvalues of the resulting Toeplitz matrix. If no column has a numerically stability above $\epsilon=0.0001$, we pick the identity matrix for our Toeplitz transformation.

\subsection{De-biasing Transformation}
Oftentimes, $\W$ is not full rank, and has a nontrivial nullspace. The bias $\b$ will have some component in the direction of this nullspace, and eliminating this component only affects the behavior of the output network $g$, and the perturabtion cannot carry on to the remainder of the sequence via $f$. Therefore, we eliminate any such component, and compensate accordingly by modifying the bias in the first affine layer of $g$. We identify the nullspaces by taking an SVD and identifying components whose singular value is less than $\epsilon=0.1$.

\subsection{Quantization Transformation}
After applying all of the previous transformations to the RNN, it is common for many of the weights to become close to zero or some other small integer. Treating this as a sign that the network is attempting to implement discrete operations using integers, we snap any weights and biases that are within a threshold $\epsilon=0.01$ of an integer, to that integer. For certain simple tasks, sometimes this allows the entire network to become quantized.

\section{Supplementary training data details} \label{extra_task_explanation}

\def\v{{\bf v}}
\def\F{{\bf F}}

Here we present additional details on the benchmark tasks marked "see text" in Table \ref{tab:tasks}:
\begin{itemize}
    \item \textbf{Div\_3/5/7:} This is a long division task for binary numbers. The input is a binary number, and the output is that binary number divided by 3, 5, or 7, respectively. The remainder is discarded. For example, we have 1000011/11=0010110 (67/3=22). The most significant bits occur first in the sequence.
    \item \textbf{Dithering:} This is a basic image color quantization task, for 1D images. We map 4-bit images to 1-bit images such that the cumulative sum of pixel brightnesses of both the original and dithered images remains as close as possible.
    \item \textbf{Newton\_Gravity:} This is an euler forward propagation technique which follows the equation $F = \text{\textit{input}} - 1, v \mapsto v + F, x \mapsto x + v$.
    \item \textbf{Newton\_Spring:} This is an euler forward propagation technique which follows the equation $F = \text{\textit{input}} - x, v \mapsto v + F, x \mapsto x + v$.
    \item \textbf{Newton\_Magnetic:} This is an euler forward propagation technique which follows the equation $F_x = \text{\textit{input}}_1 - v_y, F_y = \text{\textit{input}}_2 + v_x, \v \mapsto \v + \F, \x \mapsto \x + \v$.
\end{itemize}

% Used majority, max, min to be solvable with ReLU activations.

\section{Generated programs}\label{app:generated-programs}

\begin{comment}

\end{comment}

\definecolor{codegreen}{rgb}{0,0.6,0}
\definecolor{codegray}{rgb}{0.5,0.5,0.5}
\definecolor{codepurple}{rgb}{0.58,0,0.82}
\definecolor{backcolour}{rgb}{0.95,0.95,0.92}

\lstdefinestyle{mystyle}{
    backgroundcolor=\color{backcolour},   
    commentstyle=\color{codegreen},
    keywordstyle=\color{magenta},
    numberstyle=\tiny\color{codegray},
    stringstyle=\color{codepurple},
    basicstyle=\ttfamily\footnotesize,
    breakatwhitespace=false,         
    breaklines=true,                 
    captionpos=b,                    
    keepspaces=true,                 
    numbers=left,                    
    numbersep=5pt,                  
    showspaces=false,                
    showstringspaces=false,
    showtabs=false,                  
    tabsize=2
}

\lstset{style=mystyle}

This section includes all successfully generated Python programs.

\newcount\tmpnum
\def\storedata#1#2{\tmpnum=0 \edef\tmp{\string#1}\storedataA#2\end}
\def\storedataA#1{\advance\tmpnum by1
   \ifx\end#1\else
      \expandafter\def\csname data:\tmp:\the\tmpnum\endcsname{#1}%
      \expandafter\storedataA\fi
}
\def\getdata[#1]#2{\csname data:\string#2:#1\endcsname}

%% define tasks in \mydata %%
\storedata\mydata{
{Binary-Addition}
{Bitwise-Xor}
{Bitwise-Or}
{Bitwise-And}
{Bitwise-Not}
{Parity-Last2}
{Parity-Last3}
{Parity-All}
{Parity-Zeros}
{Sum-All}
{Sum-Last2}
{Sum-Last3}
{Sum-Last4}
{Sum-Last5}
{Sum-Last6}
{Sum-Last7}
{Current-Number}
{Prev1}
{Prev2}
{Prev3}
{Prev4}
{Prev5}
{Previous-Equals-Current}
{Diff-Last2}
{Abs-Diff}
{Abs-Current}
{Bit-Shift-Right}
{Bit-Dot-Prod-Mod2}
{Add-Mod-3}
{Newton-Freebody}
{Newton-Gravity}
{Newton-Spring}
{Dafny-Code}}

\begin{multicols}{2}
\noindent
\input{./programs/control_program_input.tex}
\end{multicols}

\subsection{Formal Verification}\label{app:dafnyverification}

The Dafny programming language is designed so that programs can be formally verified for correctness. The desired behavior of a program can be explicitly specified via preconditions, postconditions, and invariants, which are verified via automated theorem proving. These capabilities make Dafny useful in fields where correctness and safety are crucial.

We leverage Dafny's robust verification capabilities to prove the correctness of the bit addition Python program synthesized by MIPS. The bit addition Python program was first converted to Dafny, then annotated with specific assertions, preconditions, and postconditions that defined the expected behavior of the code. Each annotation in the code was then formally verified by Dafny, ensuring that under all possible valid inputs, the code's output would be consistent with the expected behavior. On line 79, we show that the algorithm found by MIPS is indeed equivalent to performing bit addition with length 10 bitvectors in Dafny. 

\input{./programs/dafny.tex}

%\end{comment}

\end{document}

%% file: tables/architecture_search_results.tex
1 & Binary\_Addition & 2 & 1 & 1 & 4 & 2 & 0 & 0 \\
2 & Base\_3\_Addition & 2 & 1 & 1 & 5 & 2 & 0 & 0 \\
3 & Base\_4\_Addition & 2 & 1 & 1 & 5 & 2 & 0 & 0 \\
4 & Base\_5\_Addition & 2 & 1 & 1 & 5 & 2 & 0 & 0 \\
5 & Base\_6\_Addition & 2 & 1 & 1 & 6 & 2 & 2.45e-09 & 2.53e-09 \\
6 & Base\_7\_Addition & 2 & 1 & 1 & 10 & 2 & 2.32e-06 & 2.31e-06 \\
7 & Bitwise\_Xor & 1 & 1 & 1 & 2 & 2 & 0 & 0 \\
8 & Bitwise\_Or & 1 & 1 & 1 & 1 & 1 & 3.03e-02 & 3.03e-02 \\
9 & Bitwise\_And & 1 & 1 & 1 & 1 & 1 & 3.03e-02 & 3.03e-02 \\
10 & Bitwise\_Not & 1 & 1 & 1 & 1 & 1 & 0 & 0 \\
11 & Parity\_Last2 & 1 & 1 & 1 & 229 & 2 & 1.68e-02 & 1.69e-02 \\
12 & Parity\_Last3 & 2 & 1 & 1 & 5 & 2 & 1.62e-04 & 1.64e-04 \\
13 & Parity\_Last4 & 3 & 1 & 1 & 29 & 2 & 3.07e-07 & 2.99e-07 \\
14 & Parity\_All & 1 & 1 & 1 & 2 & 2 & 0 & 0 \\
15 & Parity\_Zeros & 1 & 1 & 1 & 2 & 2 & 0 & 0 \\
16 & Evens\_Counter & 4 & 1 & 1 & 73 & 3 & 8.89e-05 & 8.88e-05 \\
17 & Sum\_All & 1 & 1 & 1 & 1 & 1 & 6.09e-08 & 6.13e-08 \\
18 & Sum\_Last2 & 2 & 1 & 1 & 1 & 1 & 0 & 0 \\
19 & Sum\_Last3 & 3 & 1 & 1 & 1 & 1 & 6.34e-07 & 6.35e-07 \\
20 & Sum\_Last4 & 4 & 1 & 1 & 1 & 1 & 2.10e-04 & 2.11e-04 \\
21 & Sum\_Last5 & 5 & 1 & 1 & 1 & 1 & 8.86e-03 & 8.87e-03 \\
22 & Sum\_Last6 & 6 & 1 & 1 & 1 & 1 & 1.82e-02 & 1.81e-02 \\
23 & Sum\_Last7 & 7 & 1 & 1 & 1 & 1 & 3.03e-02 & 3.01e-02 \\
24 & Current\_Number & 1 & 1 & 1 & 1 & 1 & 0 & 0 \\
25 & Prev1 & 2 & 1 & 1 & 1 & 1 & 0 & 0 \\
26 & Prev2 & 3 & 1 & 1 & 1 & 1 & 0 & 0 \\
27 & Prev3 & 4 & 1 & 1 & 1 & 1 & 0 & 0 \\
28 & Prev4 & 5 & 1 & 1 & 1 & 1 & 2.04e-07 & 2.05e-07 \\
29 & Prev5 & 6 & 1 & 1 & 1 & 1 & 6.00e-05 & 5.96e-05 \\
30 & Previous\_Equals\_Current & 2 & 1 & 1 & 5 & 2 & 6.72e-05 & 6.61e-05 \\
31 & Diff\_Last2 & 2 & 1 & 1 & 1 & 1 & 0 & 0 \\
32 & Abs\_Diff & 2 & 2 & 2 & 1 & 1 & 1.84e-07 & 1.84e-07 \\
33 & Abs\_Current & 1 & 1 & 1 & 2 & 2 & 4.51e-08 & 5.71e-08 \\
34 & Diff\_Abs\_Values & 2 & 1 & 1 & 4 & 2 & 3.15e-06 & 2.96e-06 \\
35 & Min\_Seen & 1 & 1 & 1 & 2 & 2 & 0 & 0 \\
36 & Max\_Seen & 1 & 1 & 1 & 2 & 2 & 1.46e-12 & 0 \\
37 & Majority\_0\_1 & 1 & 1 & 1 & 63 & 2 & 4.03e-03 & 4.05e-03 \\
38 & Majority\_0\_2 & 4 & 1 & 1 & 98 & 2 & 1.64e-04 & 1.71e-04 \\
39 & Majority\_0\_3 & 21 & 1 & 1 & 132 & 3 & 6.94e-05 & 6.86e-05 \\
40 & Evens\_Detector & 5 & 1 & 1 & 163 & 2 & 8.18e-04 & 8.32e-04 \\
41 & Perfect\_Square\_Detector & 48 & 1 & 1 & 100 & 2 & 1.92e-03 & 1.97e-03 \\
42 & Bit\_Palindrome & 18 & 1 & 1 & 86 & 2 & 3.81e-05 & 3.69e-05 \\
43 & Balanced\_Parenthesis & 1 & 1 & 1 & 16 & 2 & 7.44e-03 & 7.10e-03 \\
44 & Parity\_Bits\_Mod2 & 1 & 1 & 1 & 1 & 1 & 0 & 0 \\
45 & Alternating\_Last3 & 2 & 1 & 1 & 3 & 2 & 1.85e-02 & 1.87e-02 \\
46 & Alternating\_Last4 & 2 & 1 & 1 & 3 & 2 & 8.24e-06 & 8.09e-06 \\
47 & Bit\_Shift\_Right & 2 & 1 & 1 & 1 & 1 & 0 & 0 \\
48 & Bit\_Dot\_Prod\_Mod2 & 1 & 1 & 1 & 3 & 2 & 0 & 0 \\
49 & Div\_3 & 2 & 1 & 1 & 59 & 2 & 6.40e-03 & 6.43e-03 \\
50 & Div\_5 & 4 & 1 & 1 & 76 & 2 & 1.50e-04 & 1.55e-04 \\
51 & Div\_7 & 4 & 1 & 1 & 103 & 2 & 6.65e-04 & 6.63e-04 \\
52 & Add\_Mod\_3 & 1 & 1 & 1 & 149 & 2 & 1.02e-03 & 1.04e-03 \\
53 & Add\_Mod\_4 & 2 & 1 & 1 & 33 & 2 & 1.53e-04 & 1.44e-04 \\
54 & Add\_Mod\_5 & 3 & 1 & 1 & 43 & 2 & 1.02e-03 & 1.03e-03 \\
55 & Add\_Mod\_6 & 4 & 1 & 1 & 108 & 2 & 6.14e-04 & 6.12e-04 \\
56 & Add\_Mod\_7 & 4 & 1 & 1 & 199 & 2 & 3.96e-04 & 4.07e-04 \\
57 & Add\_Mod\_8 & 67 & 1 & 1 & 134 & 2 & 8.53e-04 & 8.34e-04 \\
58 & Dithering & 81 & 1 & 1 & 166 & 2 & 7.72e-04 & 7.75e-04 \\
59 & Newton\_Freebody & 2 & 1 & 1 & 1 & 1 & 2.61e-07 & 2.62e-07 \\
60 & Newton\_Gravity & 2 & 1 & 1 & 1 & 1 & 1.81e-07 & 1.87e-07 \\
61 & Newton\_Spring & 2 & 1 & 1 & 1 & 1 & 0 & 0 \\
62 & Newton\_Magnetic & 4 & 1 & 1 & 1 & 1 & 8.59e-05 & 8.60e-05 \\

%% file: programs/dafny.tex
\newcommand{\docnr}{\getdata[33]\mydata}
\docnr
\makeatletter
\edef\docnr@wosp{\docnr\space}
\newcommand{\tocfile}{}
\edef\tocfile{./programs/\expandafter\zap@space\docnr@wosp\@empty.txt}
\makeatother
\lstinputlisting[language=C++]{\tocfile}